\documentclass{article}
\usepackage{ijcai21}

\usepackage{times}
\usepackage{soul}
\usepackage{url}
\usepackage[hidelinks]{hyperref}
\usepackage[utf8]{inputenc}
\usepackage[small]{caption}
\usepackage{graphicx}
\usepackage{amsmath}
\usepackage{amsthm}
\usepackage{booktabs}
\usepackage{algorithm}
\usepackage{algorithmic}
\usepackage{stmaryrd, amssymb}

\usepackage{enumitem}
\setlist[itemize]{leftmargin=*}

\def\zero{\boldsymbol{0}}

\def\bth{\boldsymbol{\theta}}

\def\bGa{\mathbf{\Gamma}}

\def\tr{\text{tr}}

\def\A{\mathbf{A}}

\def\B{\mathbf{B}}

\def\C{\mathbf{C}}

\def\d{\mathbf{d}}
\def\D{\mathbf{D}}

\def\H{\mathbf{H}}

\def\L{\mathbf{L}}

\def\N{\mathbf{N}}

\def\bR{\mathbb{R}}
\def\s{\mathbf{s}}

\def\W{\mathbf{W}}

\def\X{\mathbf{X}}

\def\Y{\mathbf{Y}}

\def\Z{\mathbf{Z}}

\def\tX{\underline{\X}} 
\def\tY{\underline{\Y}}

\def\tZ{\underline{\Z}}

\def\tN{\underline{\N}}
\def\calP{\mathcal{P}}

\def\lnorm{\left\|}
\def\rnorm{\right\|}
\def\lrb{\left(}
\def\rrb{\right)}

\newtheorem{proposition}{Proposition}

\graphicspath{figures/}
\title{Multi-version  Tensor Completion for Time-delayed Spatio-temporal Data}

\author{
Cheng Qian$^{1,}$\thanks{Equal contribution.}\and
Nikos Kargas$^{2,*}$\and
Cao Xiao$^{1}$\and
Lucas Glass$^{1}$\and
Nicholas Sidiropoulos$^{3}$\And
Jimeng Sun$^{4}$
\affiliations
$^1$Analytics Center of Excellence, IQVIA\\
$^2$Department of Electrical and Computer Engineering, University of Minnesota Twin Cities\\
$^3$Department of Electrical and Computer Engineering, University of Virginia\\
$^4$Department of Computer Science, University of Illinois Urbana-Champaign
\emails
\{alextoqc,danicaxiao,jimeng.sun\}@gmail.com,
karga005@umn.edu,
lucas.Glass@iqvia.com,
nikos@virginia.edu
}

\begin{document}

\maketitle

\begin{abstract}
  Real-world spatio-temporal data is often incomplete or inaccurate due to various data loading delays. For example, a location-disease-time tensor of case counts can have multiple delayed updates of recent temporal slices for some locations or diseases. Recovering such missing or noisy (under-reported) elements of the input tensor can be viewed as a generalized tensor completion problem. Existing tensor completion methods usually assume that i) missing elements are randomly distributed and ii) noise for each tensor element is i.i.d. zero-mean. Both assumptions can be violated for spatio-temporal tensor data. We often observe multiple versions of the input tensor with different under-reporting noise levels. The amount of noise can be time- or location-dependent as more updates are progressively introduced to the tensor. We model such dynamic data as a multi-version tensor with an extra tensor mode capturing the data updates. We propose a low-rank tensor model to predict the updates over time. We demonstrate that our method can accurately predict the ground-truth values of many real-world tensors. We obtain up to  $27.2\%$  lower root mean-squared-error compared to the best baseline method. Finally, we extend our method to track the tensor data over time, leading to significant computational savings.
\end{abstract}

\section{Introduction}
\label{sec:intro}

Data completion is the process of estimating missing elements of an $N$-dimensional array using the observed elements and some presumed structural properties of the data, such as non-negativity and/or low rank. In many real-world spatio-temporal data, missing values are often due to delays or failures in the data acquisition or reporting process. 
Low-rank tensor factorization and completion is a popular approach to data denoising and completion, with applications ranging from image restoration~\cite{liu2012,xu2013block} to healthcare data completion~\cite{acar2011,wang2015rubik},  recommendation systems~\cite{almutairi2017,song2017}, and link prediction~\cite{lacroix2018}. Low-rank tensor factorization models have also been proposed for analyzing and making predictions based on spatio-temporal data~\cite{bahadori2014,de2017,kargas2020}.

Existing methods usually assume that missing elements are randomly distributed, and the noise for each tensor element is zero-mean i.i.d. (independent and identically distributed). However, in many real-world tensor data, such as medical claims data and public health reporting data, both assumptions are violated. For example, insurance claim processing centers receive daily data from third-party providers, where there is often a time lag between data generation and data loading. There are time delays when hospitals report COVID case counts to the government. Such data can be modeled as multiple input tensors with different time-dependent noise levels, where more recent data are more unreliable due to the data acquisition delays. Periodically, updates on tensor elements arrive to produce a new version of the tensor. Over time, we observe multiple versions of the underlying tensor. To model such multi-version tensors, we have to deal with the following challenges.
\begin{itemize}
	\item {\bf Atypical noise model:} In our context, the main source of noise is under-reporting. Thus noise is non-positive {\em and} negatively correlated in the update mode (if one update is unusually large, the next is likely to be smaller). Under-reporting also exhibits predictable patterns (e.g., holidays and weekends).  
	\item {\bf Devising efficient schemes for incremental updates:} As data updates arrive incrementally over time for both new and historical tensor slices, it is challenging to update the tensor factorization efficiently. Existing works~\cite{Sun2006-mm,Zhou2016-tm} in dynamic tensor factorization assume that the historical data slices remain unchanged, and only new slices are added.
\end{itemize}
To address these challenges, we propose a Multi-version Tensor Completion (MTC) framework. Our key contributions are as follows:
\begin{itemize}[leftmargin=*]
	\item {\bf Multi-version tensor model:} Instead of modeling the different versions of the data tensor {\em per se}, we introduce an extra tensor mode capturing the {\em data updates} across multiple versions of the data tensor. This allows us to explicitly model the update dynamics in latent space. 
	\item {\bf Online factorization} We propose a low-rank tensor model to accurately track the updates over time. To estimate the tensor values at any given time point, we perform marginalization over the extra tensor mode. We show how we can extend our approach to track the tensor data over time, leading to significant computational savings. 
	\item We compare MTC against several tensor and time series baselines on multiple real-world spatio-temporal datasets. We observe that MTC can accurately predict the ground-truth values of many real-world tensors and achieve up to  $27.2\%$  lower root mean-squared-error than the best baseline method. 
	\item Finally, we extend our method to track the tensor data over time, leading to significant computational savings. 
\end{itemize}

\section{Related Work}
CANDECOMP/PARAFAC decomposition (CPD)~\cite{Harshman1970} and Tucker decomposition~\cite{tucker1966some} have been used for imputing missing data for image restoration~\cite{liu2012,xu2013block}, grade prediction~\cite{almutairi2017}, link prediction~\cite{lacroix2018}, and healthcare data completion~\cite{acar2011,wang2015rubik}. Under certain low-rank assumptions, it is possible to recover the missing entries~\cite{yuan2016}. Tensor models have also been used for joint tracking and imputation~\cite{Sun2006-mm,sun2008incremental,nion2009adaptive,mardani2015,song2017} which enables dynamic tensor factorization with streaming data via multi-linear decomposition. Recently, several nonlinear tensor factorization models have been proposed.  NeuralCP~\cite{Liu2018} employs two neural networks where the first one predicts the entry value and the second one estimates the noise variance. COSTCO~\cite{Liu2019} is a convolutional tensor framework that models the nonlinear relationship using the local embedding features extracted by two convolutional layers. SPIDER~\cite{fang2020streaming} is a probabilistic deep tensor factorization method suitable for streaming data.

These methods assume that past slices remain unchanged and the noise in the tensor elements is i.i.d. zero-mean. These assumptions are not appropriate for modeling noise due to under-reporting, as explained earlier, which motivates our Multi-version Tensor Completion (MTC) framework.

\section{Problem Statement}
Given a spatio-temporal tensor of $I$ locations and $J$ features over time,
we first introduce two time concepts: 

\smallskip
\noindent\textbf{Generation date} (GD) is the time when data items are generated. 

\smallskip
\noindent\textbf{Loading date} (LD) is when we receive the data items. On an LD $t$, we can receive data items related to multiple locations and signals that have been generated in different GDs. 

Next, we will introduce several tensors at loading date $t$: 
the observed tensor $\tZ_t$, the ground-truth tensor $\tilde{\tZ}_t$, and the update tensor $\tX_t$ all up to time $t$.


On each LD $t$, we receive data updates $\{x_{ijs}(t)\}_{s \in \mathcal{S}_t}$ for location $i$, feature $j$ and GDs $\mathcal{S}_t$ which are earlier than $t$ i.e., $ s \leq t  ~ \forall s \in \mathcal{S}_t$. The total number of data that are generated on GD $s$ and are available on LD $t$ is computed by summing over all data received in the interval $[s,t]$, i.e., ${z_{ijs}(t)=\sum_{t'=s}^t x_{ijs}(t')}$. Stacking all $\{z_{ijs}(t)\}$ into a $3$-way tensor, yields ${\tZ_t \in\mathbb{R}_{+}^{I\times J\times S_t}}$, where $\tZ_t(i,j,s)=z_{ijs}(t)$. Tensor $\tZ_t$ contains the aggregated number of updates which have been received at time $t$. Because of delays in the data acquisition process, data corresponding to the most recent GDs are usually under-reported. Let $\widetilde{\tZ}_t$ be the ground truth tensor at time $t$. The signal model can then be described by
\begin{equation}
\tZ_{t} = \widetilde{\tZ}_{t} + \tN_{t},~t=1,2,\ldots,
\label{eq:signal_model}
\end{equation}
where $\N_{t} \in\mathbb{R}^{I\times J\times S_t}$ denotes a noise tensor. Our goal is to estimate the groundtruth tensor $\widetilde{\tZ}_t$.
We identify three main challenges in estimating $\widetilde{\tZ}_t$.

\begin{enumerate}
	\item The values corresponding to the latest GDs in $\tZ_t$, i.e., the latest frontal slabs of $\tZ_t$ are under-reported and thus very noisy.  We need to unveil the update patterns from past data and correct these highly corrupted slabs.
	
	\item The noise distribution is unknown in practice. The number of noise changes over time as more corrections to the tensor is introduced.  Also,  there are patterns in the noise. The method needs to be flexible enough so that it can capture different noise distributions. 
	
	\item The dimension corresponding to the GDs in $\tZ_{t}$ is gradually growing as $t$ increases and more data are introduced. Note that new updates arrive continuously. Therefore, the method needs to be adaptive such that it cannot only correct the under-reported numbers but also refine itself efficiently as soon as new data are received.
\end{enumerate}

\begin{figure}[h!]
	\centering
	\includegraphics[width=0.8\linewidth]{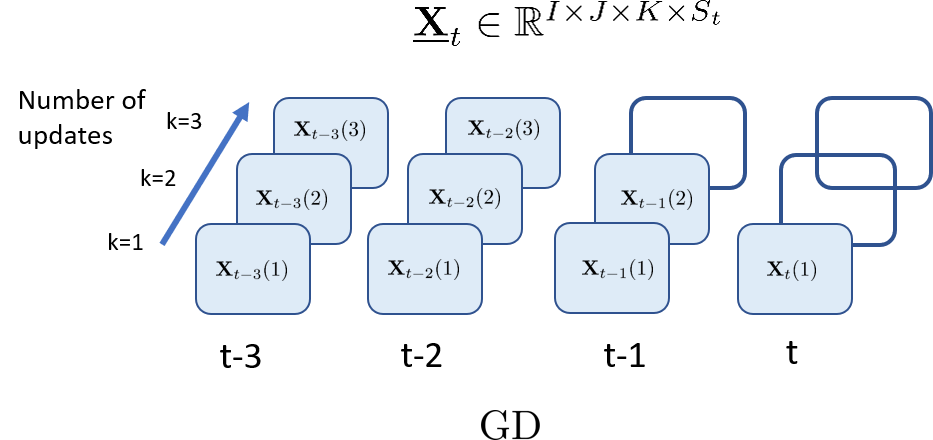}
	\caption{The updates for the data on GD $t$.}
	\label{fig:visualization}
\end{figure}

To tackle these challenges,
the key idea is to track the update tensor. We may assume that data corresponding to a given GD is updated at most $K$ times. This implies that at time $t$ we receive data for GDs $(t-K+1)$ to $t$. An example is shown in Figure~\ref{fig:visualization},  
where we assume $K=3$ and $\X_t(k)$ denotes the $k$th update of the data for $I$ location and $J$ features generated on GD $t$. Each column in Figure~\ref{fig:visualization} shows the updates for the data on GD $t$. In this example, GD $(t-3)$ has received three updates at time $t-3$, $t-2$ and $t-1$ respectively. By summing over the three updates, we have the ground truth values. 
Therefore, we have transformed the problem into an equivalent $4$-way tensor completion problem. The final estimate of the target tensor $\widetilde{\tZ}_t$  can be estimated by first imputing the missing values of $\tX_t$, and then marginalizing over the third mode 
\begin{align}\label{solver:Z}
\hat{\tZ}_t = \sum_{k=1}^K {\tX}_t(\colon, \colon, k, \colon).
\end{align}
where ${\tX}_t(\colon, \colon, k, \colon)$ is the MATLAB notation that selects data from a tensor. 

\section{Multi-version Tensor Completion (MTC)}

\noindent\textbf{Formulation}
In this work, we approximate $\tX$ using a low-rank CPD model~\cite{Harshman1970}.  For the ease of notation, w  drop the underscript $t$.
CPD expresses tensor $\tX$ as a sum of $F$ rank-$1$ tensors,
\begin{equation}
\tX =  \sum_{f=1}^F\A(:,f) \circ \B(:,f) \circ \C(:,f) \circ \D(:,f),
\end{equation}
where $\A \in\bR^{I\times F}$, $\B \in\bR^{J\times F}$,$\C \in\bR^{K\times F}$ and $\D \in\bR^{S \times F}$ are the factor matrices for location, feature, loading date (LD) and and generation date (GD), respectively. 
This kind of low-rank model is intuitive in many applications: Counties in the same state can share the same data loading schedule, and thus they have similar loading patterns. This is also true for updates data collected and sent through the same insurance company. Therefore, there is a high correlation among the data points for similar locations and features, indicating low-rank structures in the tensor. Suppose the optimal low-rank approximation of $\tX$ is $\tX \approx [\![ \A,\B,\C,\D ]\!].$

Recall that the received data can be split into two parts -- a fully observed tensor corresponding to GDs from $1$ to $(S-K+1)$ and an incomplete tensor corresponding to GDs from $(S-K+2)$ to $S$. The loss function is then expressed as
\begin{align}\label{eq:loss1}
\mathcal{F}(\bth) = \underbrace{\alpha\mathcal{F}_1(\bth)}_{\text{fully observed}} + \underbrace{(1-\alpha)\mathcal{F}_2(\bth)}_{\text{incomplete}}
\end{align}
where 
$$\begin{aligned}
	\mathcal{F}_1(\bth)=\sum_{s=1}^{S-K+1} \|\tX(:,:,:,s) - [\![ \A,\B,\C,\d_s]\!]\|_F^2, \\
\mathcal{F}_2(\bth)=\sum_{s=S-K+2}^{S}\|\calP_{\Omega_{s}}(\tX(:,:,:,s) - [\![\A,\B,\C,\d_s]\!])\|_F^2
\end{aligned}
$$
where $\|\cdot\|_F$ is the Frobenius norm, $\d_s$ denotes the $s$-th row of $\D$, $\Omega_s$ denotes the index set of the observed entries of ${\tX(:,:,:,s)}$, and $\calP_{\Omega_s}(\cdot)$ is the operation for tensor completion by only keeping the entries from index set $\Omega_s$ and zero-out the remaining entries, and $\bth = \{\A,\B,\C,\D\}$. 

In the context of spatio-temporal analysis, we would like to keep the latent factors interpretable. Towards this end, we employ a graph regularizer which regularizes the latent subspace of locations according to its graph Laplacian $\L$,  such that two locations are sufficiently close to each other in the latent subspace if they are connected in the graph and have similar loading patterns. This regularization term is
\begin{align*}
\mathcal{R}(\A) = \rho_A\tr\lrb \A^T\L\A \rrb,
\end{align*}
where $^T$ is the transpose and $\rho_A$ is the regularization parameter.
Furthermore, for the LD factor $\C$ and GD factor $\D$,
we impose smoothness regularization, i.e., 
\begin{align*}
\mathcal{R}(\C) = \rho \|\bGa \C\|_F^2,~  \mathcal{R}(\D) = \rho \|\bGa \D\|_F^2.
\end{align*}
We choose $\bGa $ to be a Toeplitz matrix holding $(-1, 2, -1)$ as its principal three diagonals and zeros elsewhere. 
$\bGa $ promotes smoothness on the latent factors so that data corresponding to adjacent time points has a smooth transition in LD and GD dimensions. 
Let $\mathcal{R}(\bth)=\mathcal{R}(\A)  + \mathcal{R}(\C) + \mathcal{R}(\D)$. We propose solving
\begin{align}\label{prob1}
\min_{\bth}~ \mathcal{F}(\bth) + \mathcal{R}(\bth),~\mathrm{s.~t.}&~\bth\geq 0,
\end{align}
where the constraint $\bth\geq 0$ is selected to ensure that the imputed values are non-negative.

\smallskip
\noindent\textbf{Initialization}
Due to the special structure of the tensor completion problem in Equation~\eqref{eq:loss1} we can obtain a good initialization for our algorithm by first solving
\begin{align*}
\min_{\A,\B,\C,\D} \mathcal{F}_1(\bth),~\mathrm{s.t.}~\A\geq 0,\B\geq 0,\C\geq 0,\d_s\geq 0,
\end{align*}
for $s=1,\ldots,S-K+1$. This is a standard tensor factorization problem which can be solved efficiently~\cite{xu2013block,fu2020computing}. By solving the above problem we obtain initial estimates of $\A,\B,\C$. Then, $\mathcal{F}_2(\bth)$ reduces to a non-negative least squares problem, i.e.,
\begin{align*}
\min_{\D} \|\calP_{\Omega_{s}}( \Vec (\tX(:,:,:,s)) - (\A\odot\B\odot\C)\d_s^T)\|_2^2,   ~\d_s\geq 0
\end{align*}
for $s=S-K+2,\ldots,S$, a convex optimization problem which can be solved optimally, where $\mathrm{vec}(\cdot)$ is the vectorization operator.

\smallskip
\noindent\textbf{Optimization}
Having obtained initial estimates for $\A,\B,\C,\D$ we turn our attention to optimization Problem~\eqref{prob1}.  A common way of tackling Problem~\eqref{prob1} is through alternating optimization. 
We introduce an auxiliary variable $\tY\in\bR^{I\times J\times K\times S}$ of the same size as $\tX$ and alternatively estimate the factor matrices while imputing the missing
data leading to an expectation maximization (EM)-like approach. 
Now let us express optimization problem~\eqref{prob1} as
\begin{equation}
\begin{aligned}
\min_{\bth,\tY}~& \mathcal{F}(\bth,\tY) + \mathcal{R}(\bth) \\
\mathrm{s.~t.}~&\bth\geq 0, \calP_{\Omega_{\s}}(\tY(:,:,:,s)) = \calP_{\Omega_{\s}}(\tX(:,:,:,s)), \\
&\forall \s=S-K+2,\ldots,S,
\end{aligned}
\label{prob:1}
\end{equation}
where
\begin{align*}
&\mathcal{F}(\bth, \tY) = \alpha\mathcal{F}_1(\bth,\tY) + (1-\alpha)\mathcal{F}_2(\bth,\tY),\\
& \mathcal{F}_1(\bth, \tY)=\sum_{s=1}^{S-K+1} \|\tY(:,:,:,s) - [\![ \A,\B,\C,\d_s]\!]\|_F^2,\\
& \mathcal{F}_2(\bth, \tY)=\sum_{s=S-K+2}^{S}\|\tY(:,:,:,s) - [\![\A,\B,\C,\d_s]\!]\|_F^2.
\end{align*}
Optimization problems~\eqref{prob1} and~\eqref{prob:1} are equivalent~\cite{xu2013block}. Problem \eqref{prob:1} can be readily handled through alternating optimization. At each iteration we fix all variables except for one. 
At $\tau$-th iteration, the subproblem with respect to $\A$ is given by
\begin{align}\label{prob:A}
\min_{\A\geq \zero}~ \mathcal{F}(\A) + \mathcal{R}(\A).
\end{align}
To proceed, we define $\bar{\tY}^{(\tau)} = \tY^{(\tau)}\times_4\W$, $\bar{\D}^{(\tau)}=\W\D^{(\tau)}$, with
$\W = \mathrm{diag}\big(\sqrt{\alpha}, \cdots,\sqrt{\alpha}, \sqrt{1-\alpha}, \cdots,\sqrt{1-\alpha}\big).$
Using the mode-$1$ unfolding matrix unfolding~\cite{Sidiropoulos2017} we equivalently express $\mathcal{F}(\A) $ as
\begin{equation}
\mathcal{F}(\A) = \left \| \bar{\Y}_{(1)}^{(\tau)} - \A(\H_A^{(\tau)})^T  \right \|_F ^2,
\end{equation}
where $\H_A^{(\tau)} =\B^{(\tau)}\odot\C^{(\tau)}\odot\bar{\D}^{(\tau)}$ and matrix $\bar{\Y}_{(1)}^{(\tau)} \in \mathbb{R}_+^{I \times JKS}$ is the mode-$1$ matrix unfolding which stores the mode-$1$ fibers of the tensor as its columns.

Let $f(\A) = \mathcal{F}(\A) + \mathcal{R}(\A)$ which can be approximated by the following
quadratic function locally around $\hat{\A}^{(\tau)}$
\begin{align}\label{quad:A1}
f(\A; \hat{\A}^{(\tau)}) =&~ f(\hat{\A}^{(\tau)}) + \left\langle\nabla_{\A}\big(f(\hat{\A}^{(\tau)})\big), \A - \hat{\A}^{(\tau)}\right\rangle \notag\\
&+ \frac{\gamma_A^{(\tau)}}{2}\lnorm\A - \hat{\A}^{(\tau)}\rnorm_F^2,
\end{align}
where $\langle\cdot\rangle$ is the inner product, $\gamma_A^{(\tau)}$ is a parameter associated with the step-size. The gradient $\nabla_{\A}\big(f(\hat{\A}^{(\tau)})\big)$  is given by $\nabla_{\A}\big(f(\hat{\A}^{(\tau)})\big) = \Big(\hat{\A}^{(\tau)}(\H_A^{(\tau)})^T - \Y_{(1)}^{(\tau)} \Big)\H_A^{(\tau)} + \lambda \L\hat{\A}^{(\tau)}$ and
\begin{align}\label{extra:A}
\hat{\A}^{(\tau)} = \A^{(\tau)} + \nu^{(\tau)}(\A^{(\tau)} - \A^{(\tau-1)})
\end{align}
denotes an extrapolated point, $\nu^{(\tau-1)}\geq 0$ is the extrapolation weight which can be updated as \cite{beck2009fast} 
$ \nu^{(\tau)} = \frac{1 - e^{(\tau)}}{e^{(\tau+1)}}$
where $e^{(0)} = 0$ and $e^{(\tau)} = \frac{1 + \sqrt{4(e^{(\tau-1)})^2 + 1}}{2}$. Now the subproblem w.r.t. $\A$ reduces to
\begin{align}\label{eq:subproblem_A}
\min_{\A\geq \zero}  f(\A; \hat{\A}^{(\tau)})
\end{align}
which admits a closed-form expression as
\begin{align}\label{solver:A}
\A^{(\tau+1)} = \A^{(\tau)} - \nabla_{\A}\big(f(\hat{\A}^{(\tau)})\big) / \gamma_A^{(\tau)}.
\end{align}
Following the above analysis, we can easily write down the updates for $\B,\C$, and $\D$, i.e.,
\begin{align}
\B^{(\tau+1)} &= \B^{(\tau)} - \nabla_{\B}\big(f(\hat{\B}^{(\tau)})\big) / \gamma_B^{(\tau)} \label{solver:B} \\
\C^{(\tau+1)} &= \C^{(\tau)} - \nabla_{\C}\big(f(\hat{\C}^{(\tau)})\big) / \gamma_C^{(\tau)} \label{solver:C} \\
\D^{(\tau+1)} &= \D^{(\tau)} - \nabla_{\D}\big(f(\hat{\D}^{(\tau)})\big) / \gamma_D^{(\tau)} \label{solver:D}
\end{align}
where $\hat{\B}^{(\tau)}, \hat{\C}^{(\tau)}$ and $\hat{\D}^{(\tau)}$ have the same definition as $\hat{\A}^{(\tau)}$ in \eqref{extra:A}~\footnote{See Supplementary 1 for the analytic expressions of the gradients in \eqref{solver:B}-\eqref{solver:D}.}.
Finally, the update for $\tY$ is given by
\begin{align}\label{solver:y}
\tY^{(\tau+1)} &= 
\calP_{\Omega^c} \left ( \hat{\tX}^{(\tau+1)} \right) + \calP_{\Omega}\left(\tX \right),
\end{align}
where $\hat{\tX}^{(\tau+1)} = [\![ \A^{(\tau+1)},\B^{(\tau+1)},\C^{(\tau+1)},\D^{(\tau+1)} ]\!]$, $\Omega$ denotes the index set of the observed entries of $\tX$ and  $\Omega^c$ is the complementary set of $\Omega$.



\smallskip
\noindent\textbf{Summary of MTC}
Note that for the factor matrices, we perform local prox-linear approximation to update them. The benefits of doing this are that each subproblem admits a closed-form solution, and the algorithm is guaranteed to converge~\cite{xu2013block}.

As an example, consider the update for $\A$. Since the original subproblem is a least squares problem, the prox-linear mapping is known to have a Lipschitz continuous gradient. The smallest Lipschitz constant is the maximum eigenvalue of $(\H_A^{(\tau)T}\H_A^{(\tau)} + \lambda \L)$. This is also true for the update of $\B,\C,\D$, where their respective smallest Lipschitz constants are the largest eigenvalues of $\H_B^{(\tau)T}\H_B^{(\tau)}$, 
$(\H_C^{(\tau)T}\H_C^{(\tau)} + \rho \bGa ^T\bGa )$ and $(\H_D^{(\tau)T}\H_D^{(\tau)} + \rho \bGa ^T\bGa )$, respectively. 
In the above, $\H_B^{(\tau)} = \A^{(\tau+1)}\odot\C^{(\tau)}\odot\bar{\D}^{(\tau)},  \H_C^{(\tau)} = \A^{(\tau+1)}\odot\B^{(\tau+1)}\odot\bar{\D}^{(\tau)} ~\text{and}~
\H_D^{(\tau)} = \A^{(\tau+1)}\odot\B^{(\tau+1)}\odot\C^{(\tau+1)}$.
Furthermore, the subproblem w.r.t. $\tY_\tau$ is also convex with closed-form solution. 
It follows from \cite{xu2013block}(Theorem 2.8):

\begin{proposition}
	MTC converges to a stationary point of \eqref{prob:1}.
\end{proposition}

The detailed steps of MTC are summarized in Algorithm 1 in the supplementary material. We note that MTC belongs to a family of block coordinate descent (BCD) algorithms. 
The computational bottleneck is the matricized tensor times Khatri-Rao product (MTTKRP) in the gradient calculation, which dominates the per-iteration complexity. This can be alleviated by exploiting sparsity~\cite{smith2015splatt}.
The overall complexity per iteration of MTC is $\mathcal{O}(IJKSF)$.

\section{Adaptive Update}
The MTC algorithm leverages all data to complete the missing values, but this is at the expense of high complexity. It requires solving the tensor completion problem exactly at any instant time $t$. To avoid restarting MTC, we propose an efficient forward-backward propagation based strategy to estimate the missing values and update all factors simultaneously. The resulting algorithm is termed MTC-online. 

Let $\tX_{t+1}\in\mathbb{R}_{+}^{I\times J\times K\times S_{t+1}}$ be obtained after appending newly observed data $\breve{\tX}_{t+1} \in \mathbb{R}_{+}^{I\times J\times K }$ in the fourth dimension, where $S_{t+1}=S_t + 1$. In this case, some of the missing values in $\tX_t$ have been completed, but new missing values are included. Therefore, the index set $\Omega_t$ changes to $\Omega_{t+1}$ and the factor matrices should be refined accordingly. Assume that the best low-rank approximation of the missing values is
\begin{align*}
\tX_{t+1} \approx   [\![ \A_{t+1}, \B_{t+1},\C_{t+1},\D_{t+1} ]\!],
\end{align*}
where the dimension of $\A_{t+1}$, $\B_{t+1}$ and $\C_{t+1}$ are the same as their counterparts obtained at time $t$ while  the number of rows in $\D_{t+1}$ is now expanded to $(S_t+1)$. 

To develop a linear complexity algorithm, it is necessary to impose a proper approximation for the new data. The choice of such an approximation has a major effect on the performance of the resulting algorithm. One such option is to assume slow variations between data from $t$ to $(t+1)$. This means that the $(i,j,k)$-th entry in $\breve{\tX}_{t+1}$ can be well approximated using the factors from the previous time step. Thus, we have
\begin{align}
\mathrm{vec}(\breve{\tX}_{t+1}) \approx (\A_t \odot\B_t \odot\C_t)\big(\D_{t+1}(S_{t+1},:)\big)^T.
\end{align}

We then employ a two-step approach to refine the factor matrices, including one forward-propagation (FP) update and one backward-propagation (BP) update, where in each propagation, only one iteration is employed to refine the factor matrices. Specifically, in the FP step, we update the row of the latent factor corresponding to the new GD. The associated optimization problem is described as
\begin{align}\label{prob:dt1}
\min_{\d \geq 0} \|  \mathrm{vec}(\breve{\tX}_{t+1}) - (\A_t \odot\B_t \odot\C_t) \d   ) \|_2^2,
\end{align}
which can be solved optimally using projected gradient descent with light-weight operations. After we obtain $\d$, we perform the BP. We first append $\d$ to $\D_t$ and then complete all missing values via
\begin{equation}
\begin{aligned}
\!\!\!\tY_{t+1}   \!=\!  \calP_{\Omega^c_{t+1}}\!(\tX_{t+1}) \!+\!   \calP_{\Omega_{t+1}}\!\big([\![ \A_t,\B_t,\C_t, [\D^T_t,\d]^T ]\!]\big)
\end{aligned}
\label{solve:Yt1}
\end{equation}
where $\Omega_{t+1}$ is the index set of the the missing values at time $(t+1)$ and $\Omega^c_{t+1}$ is its complement.
We refine all factor matrix through one iteration of the gradient updates in \eqref{solver:A}-\eqref{solver:D} in a coordinate manner.

\section{Experiments}

We evaluate the MTC's performance in terms of prediction accuracy and scalability. We use the following metrics: Root Mean Squared Error (RMSE), Mean Absolute Error (MAE), and $R^2$ score. To evaluate scalability, we record the CPU running time. All methods were trained on 2.6 GHz 6-Core Intel Core i7, 16 GB Memory, and 256 GB SSD.

\subsection{Experimental Setup}
\noindent\textbf{Data}.
We consider the following datasets for evaluation.
\begin{itemize}
	\setlength\itemsep{-0.3em}
	\item Chicago-Crime dataset~\cite{frosttdataset} includes 5,884,473 crime reports in Chicago, ranging from 2001 to 2017. We us biweekly aggregation of the reports resulting in a $77 \times 32 \times 442\times 10$ tensor of density is 0.219. These are 77 locations,  32 crime categories, 442 GDs, and 10 LDs. 
	
	\item COVID-19 dataset~\cite{dong2020} summarizes the COVID-19-19 daily reports from Johns Hopkins University. It records 3 categories, i.e., new cases, deaths, and hospitalization of 51 states in the United States between April 15, 2020, to October 31, 2020. The tensor's size is $51 \times 3 \times 200 \times 8$, and its density is 0.616, where there are 51 states, 3 features, 200 GDs, and 8 LDs.
	
	\item Patient-Claims dataset extract patient claims data from a proprietary claims database from 2018. Each data sample consists of a zip code, diagnosis (i.e., disease), date of diagnosis (i.e., GD), and date that claims loaded to the database (i.e., LD). We transform the zip codes to US counties and map the diagnosis codes to 22 diseases according to the ICD-10-CM category\footnote{\url{https://www.icd10data.com/ICD10CM/Codes}}. Finally, we count the claims for a given county, disease, GD, and LD and create a $4$-way tensor. The tensor's size is $3027 \times 22 \times 52 \times 12$ (over 41m entries) and its density is 0.534, where there are 3027 counties, 22 diseases, 52 GDs and 12 LDs.
\end{itemize}



\noindent\textbf{Evaluation Strategy}.
We consider both static and dynamic cases.  We select the first $S$ 
GDs from each dataset for the static case while setting the remaining for the dynamic case. Here, $S$  is set to $421$, $168$, and $46$ for Chicago-Crime, COVID-19, and Patient-Claims, respectively. Therefore, their corresponding numbers of GDs for the dynamic case are $21$, $32$, and $6$, respectively. In the static case, the data corresponding to the first $(S-K+1)$ GDs are fully observed, and the last $(K -1 )$ GDs have missing LDs, so the corresponding true values $\{\tilde{z}_{ijs}\}$ in $\tilde{\tZ}$ are under-reported. Our target is to predict $\{\tilde{z}_{ijs}\}$ corresponding to the last $(K-1)$ GDs.

\noindent\textbf{Baselines}.
We compare our method against low-rank tensor factorization methods, a neural network tensor completion method, and deep sequential models.
\begin{itemize}   
	\item Naive: We use the received data until $t$ to benchmark the gap between the received data $\tZ_t$  and the actual data $\widetilde{\tZ}_t$.
	
	\item SDF ($3$-way): This is a $3$-way tensor completion approach using the Structured Data Fusion (SDF) framework in Tensorlab~\cite{vervliet2016}. We complete the missing data of $\tZ_t$ and use the learned model as our estimate for $\widetilde{\tZ}_t$.
	
	\item SDF ($4$-way): We consider a variation of MTC for comparison.  This is a $4$-way tensor completion approach using SDF/Tensorlab~\cite{vervliet2016} without smoothness and Laplacian regularization. 
	
	\item COSTCO \cite{Liu2019}: This is a convolutional neural network-based model for tensor completion.
	
	\item Auto Regressive Integrated Moving Average (ARIMA): We train distinct ARIMA models for different pairs of location and attributes/signals. 
	
	\item Long Short Term Memory (LSTM) network: We train LSTM to predict the values for the subsequent GDs.
\end{itemize}
The implementation details are provided in the appendix.

\subsection{Results}

Table \ref{tab:static} summarizes the results of the static case. The Naive method performs poorly, which indicates that the time-delayed data differ significantly from the ultimate ground truth. The poor performance of SDF ($3$-way) also verifies such an observation. The $3$-way completion approach can complete the missing values of the latest slabs of $\tZ_t$ but cannot correct the under-reported cases. 
MTC has consistently lower MAE and RMSE than the other methods in the three datasets. Note that in the Patient-Claims dataset, which is the most challenging one, MTC performs the best, and its performance is followed by the SDF (4-way) algorithm. Since both MTC and SDF (4-way) are $4$-way tensor completion approaches, the performance improvement of MTC is achieved by the well-designed prox-linear gradient approach and the regularization terms. We empirically observed that COSTCO was more susceptible to overfitting and did not perform well.

\begin{table}[!ht]
	\centering
	\caption{Performance comparison in \textbf{static} case.} 
	\resizebox{1\columnwidth}{!}{
		\begin{tabular}{lrrrrrrrrr}
			\toprule
			& \multicolumn{3}{c}{Patient-Claims} & \multicolumn{3}{c}{Covid-19} & \multicolumn{3}{c}{Chicago-Crime} \\
			\cmidrule{2-10}    Method & \multicolumn{1}{l}{RMSE} & \multicolumn{1}{l}{MAE} & \multicolumn{1}{l}{$R^2$} & \multicolumn{1}{l}{RMSE} & \multicolumn{1}{l}{MAE} & \multicolumn{1}{l}{$R^2$} & \multicolumn{1}{l}{RMSE} & \multicolumn{1}{l}{MAE} & \multicolumn{1}{l}{$R^2$} \\
			\midrule
			MTC   & $\mathbf{220.4}$ & $\mathbf{29.7}$ & $\mathbf{0.997}$ & $\mathbf{74.2}$ & $\mathbf{26.0}$ & $\mathbf{0.986}$ & $\mathbf{1.42}$ & $0.57$ & $\mathbf{0.983}$ \\\midrule
			Naive  & 1113.5 & $107.2$ & $0.896$ & $290.1$ & $97.1$  & $0.559$ & $4.98$   & $1.24$   & 0.594 \\
			SDF ($3$-way) & $1149.1$ & $146.2$ & $0.905$ &   $291.9$    &  $105.0$     &   $0.551$    & $4.70$      &  $1.24$     &  $0.648$ \\
			SDF ($4$-way)  & $278.7$ & $31.5$  & $0.995$ &  $101.7$     &   $31.8$    & $0.974$       &  $1.46$     &  \textbf{0.55}     & $0.981$ \\
			COSTCO & $633.5$ & $96.4$  & $0.972$ &  $203.1$     &   $99.3$    & $0.877$      &   $2.43$    &   $0.66$    & $0.908$ \\
			ARIMA & $524.2$ & $66.5$  & $0.981$ & $283.2$ & $99.8$  & $0.780$  & $3.63$   & $1.55$   & $0.915$ \\
			LSTM  & $400.6$ & $58.5$  & $0.989$ & $343.9$ & $111.8$ & $0.736$ & $3.65$   & $1.41$   & $0.876$ \\
			\bottomrule
	\end{tabular}}
	\label{tab:static}%
\end{table}

\begin{table}[!ht]
	\footnotesize
	\centering
	\caption{Performance comparison in \textbf{dynamic} case: ($\mathrm{mean} \pm \mathrm{std}$) of each metric, calculated over all the GDs in each dataset.
	}
	\resizebox{0.95\columnwidth}{!}{
		\begin{tabular}{llll}
			\toprule
			Method & RMSE  & MAE   & $R^2$ \\
			\midrule
			\multicolumn{4}{c}{Patient-Claims dataset}\\ \midrule
			MTC   & {$\mathbf{237.8\pm40.3}$} & $32.5\pm3.0$ & \textbf{$\mathbf{0.996\pm0.001}$} \\
			MTC-online & $238.5\pm37.5$ & \textbf{$\mathbf{32.2\pm2.8}$} & \textbf{$\mathbf{0.996\pm0.001}$}
			\\\hline
			SDF (4-way) & $253.4\pm59.5$ & $34.6\pm4.7$ & $0.996\pm0.002$ \\
			Naive & $1,017.5\pm69.3$ & $99.8\pm6.3$ & $0.912\pm0.012$ \\
			SDF (3-way) & $1,052.8\pm89.1$ & $131.0\pm15.3$ & $0.904\pm0.015$ \\
			COSTCO & $580.5\pm37.2$ & $87.9\pm5.3$ & $0.977\pm0.003$ \\
			ARIMA & $553.1\pm31.5$ & $74.6\pm5.2$ & $0.979\pm0.002$ \\
			LSTM  & $692.1\pm232.2$ & $98.7\pm31.8$ & $0.952\pm0.039$ \\
			\midrule
			\multicolumn{4}{c}{Covid-19 dataset}\\ \midrule
			MTC   & $\mathbf{105.4\pm26.7}$ & $\mathbf{34.9\pm8.2}$ & $\mathbf{0.983\pm0.005}$ \\
			MTC-online & $112.2\pm23.0$ & $35.4\pm6.5$ & $0.980\pm0.009$ \\\hline
			SDF (4-way) & $147.5\pm104.4$ & $42.5\pm14.6$ & $0.974\pm0.046$ \\
			Naive & $411.4\pm98.13$ & $139.2\pm32.5$ & $0.450\pm0.123$ \\
			SDF (3-way) & $525.2\pm218.6$ & $179.0\pm61.2$ & $0.391\pm0.187$ \\
			COSTCO & $310.0\pm171.0$ & $98.7\pm35.4$ & $0.880\pm0.082$ \\
			ARIMA & $399.2\pm134.7$ & $137.0\pm33.1$ & $0.597\pm0.256$ \\
			LSTM  & $386.4\pm96.2$ & $129.5\pm19.4$ & $0.791\pm0.065$ \\
			\midrule
			\multicolumn{4}{c}{ Chicago-Crime dataset}\\ \midrule
			MTC   & $\mathbf{1.45\pm0.09}$ & $0.58\pm0.02$ & $\mathbf{0.985\pm0.002}$ \\
			MTC-online & $1.47\pm0.10$ & $0.59\pm0.02$ & $0.984\pm0.002$ \\\hline
			SDF (4-way) & $1.50\pm0.11$ & $\mathbf{0.56\pm0.02}$ & $0.984\pm0.003$ \\
			Naive & $5.46\pm0.48$ & $1.30\pm0.08$ & $0.554\pm0.078$ \\
			SDF (3-way) & $5.12\pm0.46$ & $1.30\pm0.07$ & $0.626\pm0.058$ \\
			COSTCO & $2.63\pm0.38$ & $0.73\pm0.10$ & $0.947\pm0.014$ \\
			ARIMA & $3.77\pm0.43$ & $1.42\pm0.07$ & $0.893\pm0.032$ \\
			LSTM  & $4.10\pm0.50$ & $1.43\pm0.07$ & $0.848\pm0.034$ \\
			\bottomrule
	\end{tabular}}
	\label{tab:dynamic}%
\end{table}%

\noindent{\bf Accuracy} In Table~\ref{tab:dynamic}, we calculate the RMSE, MAE, and $R^2$ for each method in the dynamic case. The values were averaged over all GDs in each dataset. It can be seen that the proposed methods achieve the best RMSE and $R^2$ in the three datasets. We observed that COSTCO could easily overfit the data, so it did not perform well. ARIMA, LSTM, and SDF (3-way) did not use the LD information, thus failing to achieve satisfactory results. This indicated that using an additional mode -- LD can help solve the data delayed issue in spatio-temporal tensors. In the Patient-Claims dataset, MTC-online has a smaller MAE than MTC. In all other cases, MTC has slightly better performance than MTC-online. However, this is at the expense of high complexity, since MTC needs to be restarted once new data received. This is true for the other baselines as well. On the other hand, MTC-online updates all factors in an inexact way with a single proximal gradient update, resulting in very low-complexity.

\begin{figure}[t]
	\centering
	\includegraphics[width=0.6\linewidth,height=0.4\linewidth]{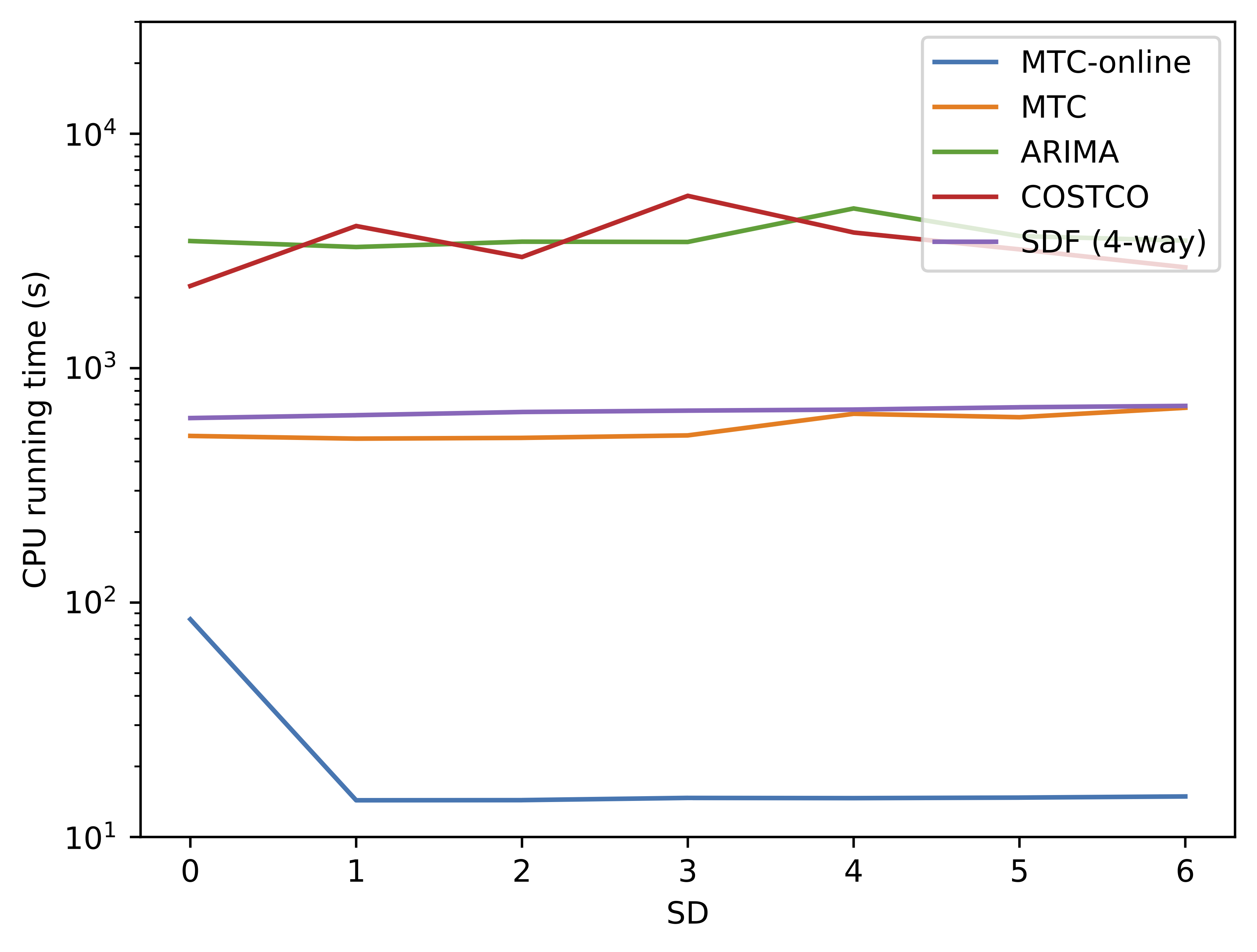}
	\vspace*{-0.05in}
	\caption{Complexity comparison. } 
	\label{fig:time}
\end{figure}

\noindent{\bf Scalability} Figure~\ref{fig:time} shows the CPU running time of MTC-online, MTC, ARIMA, and SDF (4-way) on the Patient-Claims dataset. We did not include LSTM for comparison because it had a much higher time complexity for training. Figure~\ref{fig:time} shows COSTCO and ARIMA has the highest complexity -- the CPU time for both methods was close to one hour. ARIMA is slow we need to train a separate model for each location. SDF (4-way) and MTC have similar running time. 
MTC-online is an order of magnitude faster than the best baseline due to the approximate gradient updates. 

\begin{figure}[t]
	\centering
	\includegraphics[width=.85\linewidth,height=0.55\linewidth]{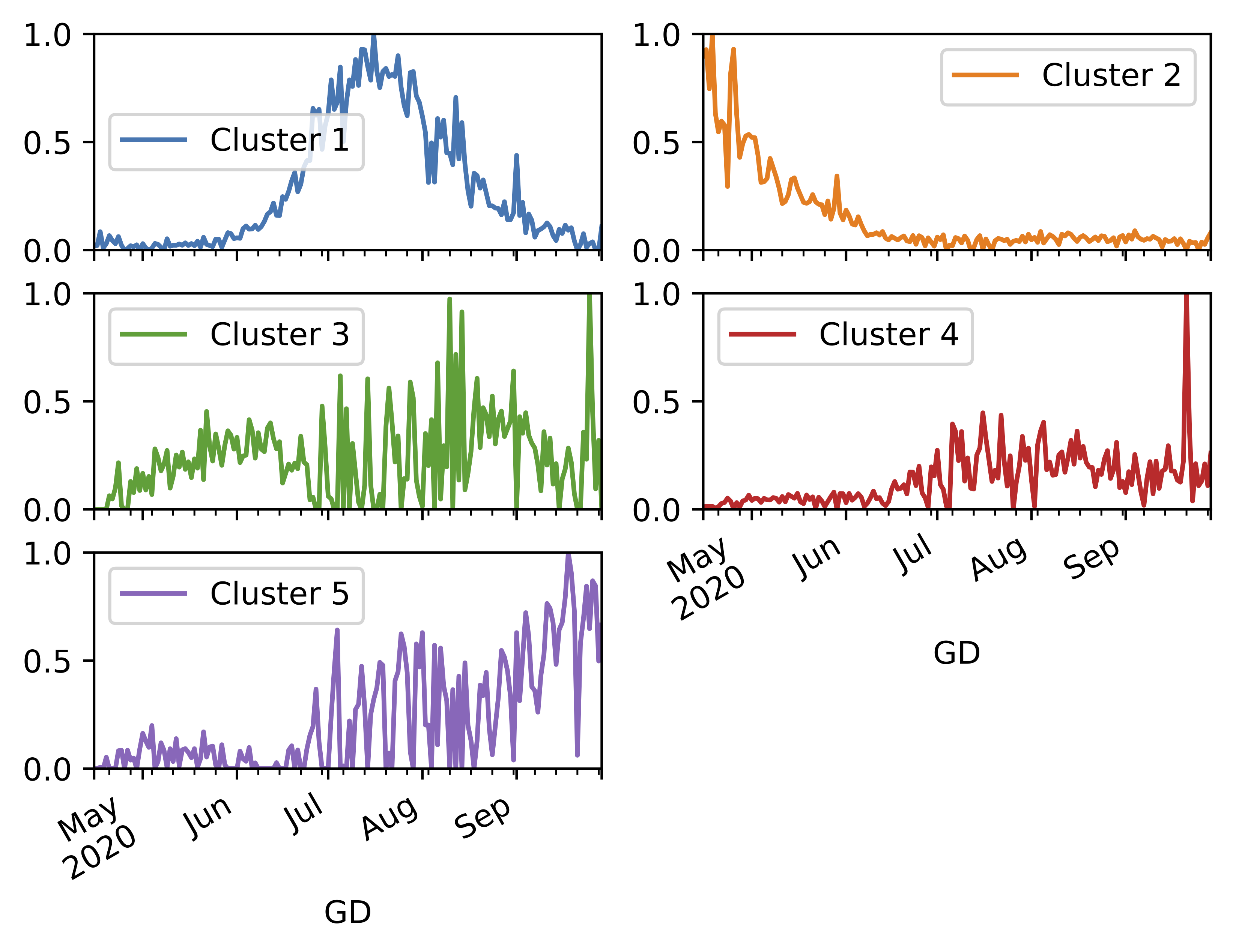}
	\caption{Latent components in GD mode of Covid-19 dataset.} 
	\label{fig:sd_jhu}
\end{figure}

\noindent{\bf Model interpretation} Figure~\ref{fig:sd_jhu} shows the latent components corresponding of the GD mode in the COVID-19 dataset. 
Figure~\ref{fig:sd_jhu} shows that Cluster 2 consisted of the states hit by Covid-19 in the first wave in the United States. The Cluster-1 had about two months delay, exactly for the second wave, including FL, CA, TX, AZ, etc. 

The five clusters have similar exponential decay trend. Interestingly, Cluster 4 had the lowest loading speed, meaning that data corresponding to this cluster took a longer time to be received than those in the other components. Four out of the top five counties in Cluster 4 were from California, where the laws on processing/collecting claims data are more strict than the other states.
More results are reported in the supplement.

\section{Conclusion}
This paper studies the time-delayed spatio-temporal tensor data, which corresponds to many real-world data aggregation applications. 
We formulated the problem as a multi-version tensor completion (MTC) problem by introducing an extra mode to capture the data updates. The final estimate of the target tensor is estimated by first imputing the missing values and then marginalizing over the extra mode. We proposed static and online version of MTC algorithms to tackle this problem. The proposed methods employ a graph regularization to reserve the location correlations in latent subspace and smoothness. The experimental results on several real datasets have demonstrated the advantages of the proposed methods.

\clearpage
\bibliographystyle{named}
\bibliography{ijcai21}

\end{document}